\documentclass[conference]{IEEEtran}
\usepackage[utf8]{inputenc}
\usepackage[T1]{fontenc}
\usepackage{multirow}

\usepackage{amsmath}
\usepackage{amssymb}
\usepackage{graphicx}
\usepackage{xcolor}
\usepackage{colortbl}
\usepackage{epsfig}
\usepackage{esvect}

\newcommand{\PM}[1]{P_{#1}}
\newcommand{\SM}[1]{S_{#1}}
\newcommand{\PstarM}[1]{P^{\star}_{#1}}
\newcommand{\SstarM}[1]{S^{\star}_{#1}}
\newcommand{\ILS}[3]{ILS_{#1}^#3(#2)}
\newcommand{\ILSs}[2]{ILS_{#1}(#2)}

{}
{}
{}

\newcommand{\trp}[2]{\paths{#1}{1}{#2}}{}
\newcommand{\paths}[3]{p_{#1,\vv{q_{#2} q_{#3}}}}{}
\newcommand{\fs}[1]{f^k_{#1}}

{}

\newcommand{\delt}[3]{\delta_{#1,\vv{q_{#2} q_{#3}}}}{}
\newcommand{\Sp}{S^{+}}{}
\newcommand{\Sm}{S^{-}}{}
{}
{}

\title{Taking advantage of a very simple property to efficiently infer NFAs}
\author{
\IEEEauthorblockN{Tomasz Jastrz\k{a}b}
\IEEEauthorblockA{\emph{Silesian University of Technology}
\\Poland\\
Email: Tomasz.Jastrzab@polsl.pl}
\and
\IEEEauthorblockN{Fr{\'{e}}d{\'{e}}ric Lardeux}
\IEEEauthorblockA{\emph{University of Angers} 
\\France\\
Email: Frederic.Lardeux@univ-angers.fr}
\and
\IEEEauthorblockN{Eric Monfroy}
\IEEEauthorblockA{\emph{University of Angers} 
\\France\\
Email: Eric.Monfroy@univ-angers.fr}
}

\date{July 2021}

\begin{document}

\maketitle

\begin{abstract}
Grammatical inference consists in learning a formal grammar as a finite state machine or as a set of rewrite rules. In this paper, we are concerned with inferring Nondeterministic Finite Automata (NFA) that must accept some words, and reject some other words from  a given sample. This problem can naturally be modeled in SAT. The standard model being enormous, some models based on prefixes, suffixes, and hybrids were designed to generate smaller SAT instances. 

There is a very simple and obvious property that says: if there is an NFA of size k for a given sample, there is also an NFA of size k+1. We first strengthen this property by adding some characteristics to the NFA of size k+1. Hence, we can use this property to tighten the bounds of the size of the minimal NFA for a given sample. We then propose simplified and refined models for NFA of size k+1 that are smaller than the initial models for NFA of size k. We also propose a reduction algorithm to build an NFA of size k from a specific NFA of size k+1. Finally, we validate our proposition with some experimentation that shows the efficiency of our approach.
\end{abstract}

\begin{IEEEkeywords}
grammatical inference, nondeterministic automata, SAT models
\end{IEEEkeywords}

\section{Introduction}
\label{sec:introduction}

Grammatical inference~\cite{ColinBook} consists in studying and designing methods for learning formal grammars (as automata or production rules) from a given sample of words. It is useful in numerous applications, e.g., compiler design, bioinformatics, pattern recognition, machine learning, etc. 

Let $S=\Sp \cup \Sm$ be a sample of words, made of positive words (the set $\Sp$) and negative words (the set $\Sm$): words from $\Sp$ must be elements of the language, and words from $\Sm$ must not. The \textbf{problem we tackle} is to learn a finite automaton which accepts words of $\Sp$ and rejects words of $\Sm$. The complexity of such problems is related to the number of states of the automaton, i.e., its size. A deterministic automaton (DFA) for a given language can be (even exponentially) larger than a nondeterministic automaton (NFA) for the same language. Thus, similarly to most of the works on automata inference, we focus here on NFAs. An NFA is represented by a 5-tuple $(Q , \Sigma , \Delta, q_1, F )$ where $Q$ is a finite set of states, the alphabet $\Sigma$ is a finite set of symbols, the transition function $\Delta  : Q \times \Sigma \rightarrow 2^Q$ associates a set of states to a given state and a given symbol, $q_1 \in Q$ is the initial state, and $F \subseteq Q$ is the set of final states. Let $k$\_NFA denote an NFA with $k$ states.
The generic problem consists in minimizing $k$. However, since most of the techniques are based on a Boolean model, the problem is simplified to learning a $k$\_NFA, $k$ being given. To minimize $k$, some lower and upper bounds are determined. For example, an upper bound is given by the size of the prefix tree acceptor (PTA). Then, some algorithms can be used to push up (resp. push down) the lower bound (resp. the upper bound) to find the smallest $k$ (see e.g.,~\cite{DBLP:journals/fuin/JastrzabCW21}).

The problem has already been tackled from several perspectives (see e.g., \cite{WieczorekBook} for a wide panel of NFA inference techniques). Ad-hoc algorithms, such as \textit{DeLeTe2}~\cite{delete2}, are based on merging states from the PTA. More recently, a new family of algorithms for regular languages inference was given in~\cite{DBLP:conf/wia/PargaGR06}. 
Some approaches are based on metaheuristics, such as in~\cite{tomita82} where hill-climbing is applied in the context of regular languages, or~\cite{DBLP:conf/icgi/Dupont94} which is based on a genetic algorithm.
In contrast to metaheuristics, complete solvers are always able to find a solution if there exists one, to prove the unsatisfiablility of the problem, and to find the global optimum in case of optimization problems. Generally, the problem is modeled as a Constraint Satisfaction Problem (CSP~\cite{Rossi2006}). For example, in~\cite{WieczorekBook}, an Integer Non-Linear Programming (INLP) formulation of the problem is given. Parallel solvers for minimizing the inferred NFA size  are presented in~\cite{jastrzab2016,jastrzab2017}. The author of~\cite{jastrzab2019} proposes two strategies, based on variable ordering, for solving the CSP formulation of the problem. Reference~\cite{DBLP:journals/fuin/JastrzabCW21} proposes a parallel approach for solving the \emph{optimization} variant of the problem.

In this paper, we do not want to improve a specific solver or to design an ad-hoc solver. On the contrary, we try to \textbf{improve the SAT models} starting from the models of~\cite{ICTAI2021}. More specifically, we focus on the results that can be obtained using a simple property: for a given sample $S$, if there is a $k$\_NFA, there is also a $(k+1)$\_NFA. This is obvious, for example by adding a new state with no incoming transition. This property can be used to find a new upper-bound and raise the lower bound (in case of UNSAT answer): this is easy to do using previous models,
but this is useless since it is more complicated to infer a $(k+1)$\_NFA than a $k$\_NFA. However, we can refine this property by requesting more characteristics of the $(k+1)$\_NFA. Assume that $\lambda \not \in S$. By design, we can request the $(k+1)$\_NFA to have a single final state without outgoing transitions, and that each incoming transition to the final state is a duplication of a transition of the $k$\_NFA; we call such an NFA a $(k+1)^\star$\_NFA. Moreover, we \textbf{simplify previous models}, which is even more interesting for the suffix and hybrid models that pass from ${\mathcal{O}}(\sigma \cdot k^3)$  clauses to ${\mathcal{O}}(\sigma \cdot k^2)$ with $\sigma=\Sigma_{w\in S}|w|$. Hence, it becomes less complicated to compute a $(k+1)^\star$\_NFA than a $k$\_NFA. The $(k+1)^\star$\_NFA can thus be used to reduce the bounds on $k$\_NFA. Finally, we \textbf{propose a linear-time algorithm} to reduce a $(k+1)^\star$\_NFA to a $k$\_NFA. The experimental results for different models are promising and show the advantage of our approach.

The paper is organized as follows. In Sect.~\ref{sec_models} we present how current models are obtained. In Sect.~\ref{sec_kpomodel}, we first present the property leading from a $k$\_NFA to a $(k+1)$\_NFA. We then show how to improve the models to infer $(k+1)^\star$\_NFA. Section~\ref{sec_algo} presents our reduction algorithm. In Sect. \ref{sec-expe} we report the experimental results and we conclude in Sect.~\ref{sec_conclusion}. 
\section{Models for $k$\_NFA inference}
\label{sec_models}
The models we now present can be used directly to infer $k$\_NFA or $(k+1)$\_NFA.
Let $\Sigma=\{a_1,\ldots,a_n\}$ be an alphabet of $n$ symbols. A sample $S=\Sp \cup \Sm$ is given by a set $\Sp$ of words from $\Sigma^{*}$ that must be accepted, and a set $\Sm$ of words that must be rejected. Let $K$ be a set of integers, $K=\{1, \ldots,k\}$. We will consider the following variables in our models:
\begin{itemize}
 \item $k$, an integer, the size of the NFA we want to learn,
 \item a set of $k$ Boolean variables $F=\{f_1, \ldots, f_k\}$  determining whether state $q_i$ is final or not,
 \item and $\Delta=\{\delt{a}{i}{j}| a \in \Sigma \textrm{~and~} (i,j) \in K^2\}$, a set of $nk^2$ Boolean variables representing the transitions from state $q_i$ to state $q_j$ with the symbol $a \in \Sigma$, for each $i$, $j$, and $a$.
\end{itemize} 

The path $i_1, i_2, \ldots, i_{m+1}$ for a word $w=a_1 \ldots  a_m$ exists \emph{iff} $d=\delt{a_1}{i_1}{i_2} \wedge  \ldots \wedge \delt{a_m}{i_{m}}{i_{m+1}}$ is true. We call $d$ a c\_path. Let $Pref(S)=\cup_{w \in S}Pref(w)$, with $Pref(w)$ the set of non-empty prefixes of word $w$. Similarly, $Suf(S)$ is the set of all non-empty suffixes of the words in $S$.

We skip the direct model (see \cite{DBLP:journals/fuin/JastrzabCW21,meta2021}) which has a bad complexity and does not behave well in practice: the space complexity is in ${\mathcal{O}}(|\Sp|\cdot(|\omega_{+}|+1)\cdot k^{|\omega_{+}|})$ clauses, and ${\mathcal{O}}(|\Sp|\cdot k^{|\omega_{+}|})$ variables with $\omega_{+}$ the longest word of~$\Sp$.

\smallskip

\noindent
\textbf{Prefix Model ($\PM{k}$)}:
For each $w \in Pref(S)$, we consider $\trp{w}{i}$,  a Boolean variable,  determining the existence of a c\_path for $w$ from $q_1$ to $q_i$. The constraints are:
 \begin{IEEEeqnarray}{c}
 (\lambda \in \Sp~\longrightarrow~f_1) ~~\wedge~~
 (\lambda \in \Sm~\longrightarrow~\neg f_1) \label{lambda} \\
 \bigvee_{i \in K} \delt{a}{1}{i} \leftrightarrow \trp{a}{i} \label{pref1} \\
 \bigwedge_{i \in K} (\trp{w}{i} \leftrightarrow (\bigvee_{j \in K} \trp{v}{j} \wedge  \delt{a}{j}{i})) \label{pref2} \\
 \bigvee_{i \in K} \trp{w}{i} \wedge f_i  \label{pref3} \\
 \bigwedge_{i \in K} (\neg \trp{w}{i} \vee \neg f_i)  \label{pref4}
 \end{IEEEeqnarray} 
 where Constraint~(\ref{lambda}) considers $\lambda$; 
 Constraint~(\ref{pref1}) specifies a c\_path of size 1 for each $a \in Pref(S) \cap \Sigma$;  
 for each $w=va$ with $w, v\in Pref(S)$, $a \in \Sigma$, Constraint~(\ref{pref2}) completes a path;
 for each word $w \in \Sp \setminus \{\lambda\}$, Constraint~(\ref{pref3}) specifies that positive words must end in a final state;
 and for each word $w \in \Sm \setminus \{\lambda\}$, Constraint~(\ref{pref4}) specifies than if there is a path for a negative word it must end in a non-final state.
 
The prefix model is defined by:
$$\PM{k} = (\ref{lambda}) \bigwedge_{w \in Pref(S)} \big((\ref{pref1}) \wedge (\ref{pref2})\big) \bigwedge_{w \in \Sp} (\ref{pref3})  \bigwedge_{w \in \Sm} (\ref{pref4})$$.

After transformations, $\PM{k}$ is converted into CNF, and its space complexity is in ${\mathcal{O}}(\sigma \cdot k^2)$ variables, and ${\mathcal{O}}(\sigma \cdot k^2)$  clauses with $\sigma=\Sigma_{w\in S}|w|$. See~\cite{meta2021} for details.

\smallskip

\noindent \textbf{Suffix Model  ($\SM{k}$)}: The construction starts from any state and terminates in state $q_1$. 
For each $w \in Suf(S)$, a Boolean variable $\paths{w}{i}{j}$ determines the existence of a c\_path for $w$:
 \begin{IEEEeqnarray}{c}
 \bigvee_{(i,j) \in K^2} \delt{a}{i}{j} \leftrightarrow \paths{a}{i}{j} \label{suf1} \\
 \bigwedge_{(i,j) \in K^2} (\paths{w}{i}{j} \leftrightarrow 
 (\bigvee_{l \in K} \delt{a}{i}{l}  \wedge  \paths{v}{l}{j})) \label{suf2}
 \end{IEEEeqnarray}
where Constraint~(\ref{suf1}) is used for suffixes of size 1, and Constraint~(\ref{suf2}) for longer ones. The model is:
$$\SM{k} = (\ref{lambda}) \bigwedge_{w \in Suf(S)} \big((\ref{suf1}) \wedge (\ref{suf2})\big) \bigwedge_{w \in \Sp} (\ref{pref3})  \bigwedge_{w \in \Sm} (\ref{pref4})$$

Although similar to $\PM{k}$, the $\SM{k}$ models are in ${\mathcal{O}}(\sigma \cdot k^3)$ variables,  and in ${\mathcal{O}}(\sigma \cdot k^3)$ clauses~\cite{meta2021}.

\smallskip

\noindent  \textbf{Hybrid Models}: Each word $w \in S$ is split into a prefix and a suffix $w=uv$ to obtain two samples $S_u=S_{u}^{+} \cup S_{u}^{-}$ with $S_{u}^{+}=\{u ~|~ \exists w \in \Sp, w=uv\}$ and 
$S_u^{-}=\{u ~|~ \exists w \in \Sm, w=uv\}$, and  
$S_v=S_{v}^{+} \cup S_{v}^{-}$ with 
$S_v^{+}=\{v ~|~ \exists w \in \Sp, w=uv\}$ and 
$S_v^{-}=\{v ~|~ \exists w \in \Sm, w=uv\}$. 
Then, Constraints~(\ref{pref1}), (\ref{pref2}) are used for prefixes of $S_u$, and (\ref{suf1}), (\ref{suf2}) for suffixes of $S_v$. Finally, for each $w=uv$, clauses generated for $u$ are linked to clauses generated for $v$:
\begin{IEEEeqnarray}{c}
\bigvee_{(i,j) \in K^2} \paths{u}{1}{j} \wedge \paths{v}{j}{i} \wedge f_i \label{hyb1}\\
\bigwedge_{(i,j) \in K^2} (\neg \paths{u}{1}{j} \vee \neg \paths{v}{j}{i} \vee \neg f_i)\label{hyb2}
\end{IEEEeqnarray}
Constraints~(\ref{hyb1}) are used for words from $\Sp$ and Constraints~(\ref{hyb2}) for words from $\Sm$.

Efficient decomposition of each word into a prefix and a suffix is crucial. In~\cite{ICTAI2021} and~\cite{ola2021}, we proposed various decomposition strategies. Here, we consider three of them:
\begin{itemize}
 \item $\ILS{k}{init}{f}$ is based on an Iterated Local Search (ILS)~\cite{Stutzle2018} with the fitness $f$ defined as $f(S_u,S_v)=|Pref(S_u)|+k \cdot |Suf(S_v)|$ for optimizing the hybrid model. The search starts with a split $init$ for each word. At each iteration, the best split $w=uv$ is found for the word $w$ selected randomly with a roulette wheel selection based on the weights of words defined by $\textrm{weight}_w = 75\% / |S| + 25 \% \cdot |w| / (\sum_{w_i \in S} |w_i|)$. The number of iterations is given and diversification is ensured with word selection.
 
 \item the \textbf{Best suffix model  ($\SstarM{k}$)} optimizes constructions of the suffixes by ordering the set $Suf(S)$. Let $
\Omega(u) = \{ w \in S ~|~ u \in Suf(w)\}$ be the set of words accepting $u$ as a suffix. Then, $u_1 \succcurlyeq u_2$ \emph{iff} $|u_1|\cdot|\Omega(u_1)| \geq |u_2|\cdot|\Omega(u_2)|$. The set of best suffixes is composed of the best suffixes (w.r.t. to $\succcurlyeq$) that cover $S$ (see~\cite{ICTAI2021} for more details).

 \item
 the \textbf{Best prefix model  ($\PstarM{k}$)} is built in a similar way as the Best suffix model, starting with a selection of the best prefixes~\cite{ICTAI2021}. 
\end{itemize}

\section{Refined models for $(k+1)$\_NFA}
\label{sec_kpomodel}
We now consider $\lambda \not \in \Sp$.

\subsection{Building a $(k+1)$\_NFA from a $k$\_NFA} From a $k$\_NFA, we can build a $(k+1)$\_NFA with a single final state without outgoing transitions, and such that each incoming transition to the final state is a duplication of a transition of the $k$\_NFA. 

Let $A=(Q^A,\Sigma, \Delta^A, q_1, F)$ be a $k$\_NFA. Then, there always exists a $(k+1)^\star$\_NFA $A'=(Q^{A'},\Sigma, \Delta^{A'}, q_1, F^{A'})$ such that $Q^{A'} = Q^A\cup\{q_{k+1}\}$, $F^{A'} = \{q_{k+1}\}$ and $\Delta^{A'}$: 
 $$
 \begin{array}{l}
 \forall_{i,j \in (Q^A)^2} \ \delt{a}{i}{j}^A \leftrightarrow \delt{a}{i}{j}^{A'}\\[2mm]
 \forall_{i \in Q^A, j \in F} \ \delt{a}{i}{j}^A \leftrightarrow \delt{a}{i}{k+1}^{A'}
 \end{array}
 $$
 
\noindent Sketch of the proof:
\begin{enumerate}
 \item For each word $w \in \Sp \setminus \{\lambda\}$:\\
 Let $w=va$ with $v \in Pref(S)$ and $a \in \Sigma$. Then $v$ is recognized by $A$ and can finish in several states $T \subseteq Q$ (not necessary final states). As $w \in \Sp$, at least one transition $\delt{a}{i}{j}^A$ with $i \in T$ and $j \in F$ exists. By the rules of transitions creation, $\delt{a}{i}{k+1}^{A'}$ exists and so word $w$ is recognized by $A'$.
 
 \item For each word $w \in \Sm \setminus \{\lambda\}$:\\
 Let $w=va$ with $v \in Pref(S)$ and $a \in \Sigma$. There may be a c\_path for $v$ in $A$ that terminates in states $T \subseteq Q$. As $w \in \Sm$, if $\delt{v}{1}{j}^A$ exists then $\delt{a}{j}{i}^A$ with $j \in T$, $i \in F$ does not exist. Thus, $\delt{a}{1}{k+1}^{A'}$ is not created and $w$ is then rejected by $A'$.
 
\end{enumerate}
\noindent
Note that if $\lambda \in \Sp$, we can do a very similar construction by considering $q_{k+1}$ and $q_1$ as both final. Then, the construction is identical with only some few more disjunctions between $q_1$ and $q_{k+1}$.

\smallskip

\begin{figure}[!ht]
 \centering
 \begin{minipage}{0.4\columnwidth}
 \includegraphics[width=\textwidth]{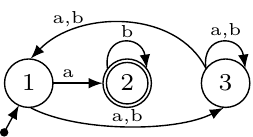}
 \end{minipage}
 \ \ 
 \begin{minipage}{0.5\columnwidth}
 \includegraphics[width=\textwidth]{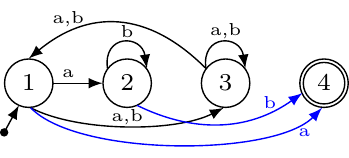}
 \end{minipage}
 \caption{Example $k$\_NFA (left) and the corresponding $(k+1)^\star$\_NFA for sample $S = (\{a, ab, abba, baa\}, \{aab, b, ba, bab\})$}
 \label{fig:examplea}
\end{figure}

\noindent\textbf{Example 1}:
Consider a sample $S = (\Sp, \Sm)$, such that $\Sp = \{a, ab, abba, baa\}$ and $\Sm = \{aab, b, ba, bab\}$. An example minimal $k$\_NFA and the corresponding $(k+1)^\star$\_NFA are shown in Fig.~\ref{fig:examplea}, with the additional transitions marked in blue. The transition $\delt{a}{1}{4}^{A'}$ results from the transition $\delt{a}{1}{2}^A$, while the transition $\delt{b}{2}{4}^{A'}$ results from the transition $\delt{b}{2}{2}^A$.

\begin{figure}[!t]
 \centering
 \begin{minipage}{0.4\columnwidth}
 \includegraphics[width=\textwidth]{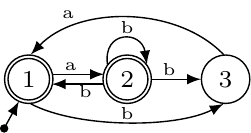}
 \end{minipage}
 \ \ 
 \begin{minipage}{0.5\columnwidth}
 \includegraphics[width=\textwidth]{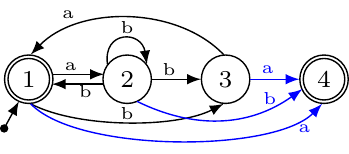}
 \end{minipage}
 \caption{Example $k$\_NFA (left) and the corresponding $(k+1)^\star$\_NFA for sample $S = (\{\lambda, ab, abba, baa\}, \{aa, aab, b, bab\})$}
 \label{fig:exampleb}
\end{figure}

\smallskip
\noindent\textbf{Example 2}:
Consider now a different sample~$S$, where $\Sp = \{\lambda, ab, abba, baa\}$, and $\Sm = \{aa, aab, b, bab\}$. The $k$\_NFA and its corresponding $(k+1)^\star$\_NFA are shown in Fig.~\ref{fig:exampleb}. The main differences with respect to the previous example are that state $q_1$ remains final in the $(k+1)^\star$\_NFA due to $\lambda \in \Sp$, and that state $q_4$ is not the final state for $\lambda$, since we assume the NFA has no $\lambda$-transitions. Note also that words $ab$ and $abba$ are accepted in both state $q_1$ and $q_4$. So, the only word that requires state $q_4$ for being accepted is $baa$. This is in contrast to Example~1, in which all the words in $\Sp$ were only accepted in state $q_4$.

\subsection{Nested NFA}
We now constrain more the $(k+1)$\_NFA in order to obtain fewer candidates.
Consider:
\begin{itemize}
 \item $F_1$: the set of $(k+1)$\_NFA for $S$
 \item $F_2$: the set of $(k+1)$\_NFA for $S$ with only one final state
 \item $F_3$: the set of $(k+1)$\_NFA for $S$ with only one final state $q_{k+1}$ and no outgoing transition from the final state 
 \item $F_4$: the set of $(k+1)$\_NFA for $S$ with only one final state $q_{k+1}$ and no outgoing transition from the final state, and with the property:
 \[
 \forall u \in Suf(\Sp)\cap\Sigma, \forall i \in K,  \delt{u}{i}{k+1} \longrightarrow \bigvee_{j \in K} \delt{u}{i}{j}
 \]

  which means that each incoming transition from $q_i$ to $q_{k+1}$ has a "clone" from $q_i$ to at least one $q_j$ which we expect to be final in the $k$\_NFA. 
  
 \item $A'$ is a $(k+1)^\star$\_NFA built by the algorithm presented before.
\end{itemize}
Then, we have that $A' \in F_4 \subseteq F_3 \subseteq F_2 \subseteq F_1$. It is thus interesting to be as close as possible to $A'$: this removes symmetries, and the reduction algorithm, from $(k+1)^\star$\_NFA to $k$\_NFA shown in Sect.~\ref{sec_algo} succeeds more frequently.

\smallskip
\noindent\textbf{Example 3}: To illustrate the importance of the $(k+1)^\star$\_NFA, let us consider the following scenario. Assume that we know that there is no NFA of size $k_1$ and that there exists an NFA of size $k_2$, $k_2 > (k_1+2)$. Now, if we find a $(k+1)^\star$\_NFA of size $k_3$, such that $k_2 \ge k_3 > (k_1+1)$, then we know that there exists an NFA of size $k_3-1$, which means that we lowered the upper bound on the size of the minimal NFA. If we also find there exists no $(k+1)^\star$\_NFA of size $k_4$ such that $k_2 > k_4 > (k_1 + 1)$, then we know there is no NFA of size $k_4 - 1$. Hence, we raised the lower bound on the size of the minimal NFA. So, we know now that the size of minimal NFA $k_{\min}$ satisfies $k_1 + 1 < k_4 \le k_{\min} \le k_3 - 1 < k_2$. Clearly, if $k_2 = k_1 + 2$, we can find either $k_3$ or $k_4$ but not both. However, it still proves the usefulness of the $(k+1)^\star$\_NFA.

\subsection{Refining and simplifying $(k+1)$\_NFA models}
\label{sec_ref}
We cannot model exactly $A'$ as described before, but we can generate $(k+1)^\star$\_NFA. The idea here is thus to refine and simplify models to obtain $(k+1)^\star$\_NFA. In what follows, we consider $\lambda \notin \Sp$.

Since we consider only one final state, $q_{k+1}$, we can omit the $f_i$ Boolean variables, and Constraints~(\ref{pref2}) and (\ref{pref3}) simplify to:
 \begin{equation}
 \trp{w}{k+1} \label{pref1s}
 \end{equation}
 for each word $w \in \Sp$.
This removes ${\mathcal{O}}(2k\cdot|\Sp|)$ constraints and ${\mathcal{O}}(k\cdot|\Sp|)$ auxiliary variables for Tseitin transformation of (\ref{pref2}), and ${\mathcal{O}}((k-1)\cdot|\Sp|)$ clauses in Constraints~(\ref{pref3}).


Constraint~(\ref{pref4}) can also be replaced with:
\begin{equation}
\neg \paths{w}{1}{k+1} \label{pref4s}
\end{equation}
for each word of $\Sm$. This replaces $k\cdot|\Sm|$ binary clauses with $|\Sm|$ unit clauses.


These simplifications are worth with the prefix and suffix models. Moreover, for the suffix model, Constraint~(\ref{suf1}) can be simplified to:
\begin{equation}
\bigvee_{i \in K} \delt{a}{i}{k+1}
\leftrightarrow 
\paths{a}{i}{k+1} \label{suf1s}
\end{equation}
and (\ref{suf2}) to:
\begin{equation}
\bigwedge_{i \in K} (\paths{w}{i}{k+1} \leftrightarrow 
 (\bigvee_{j \in K} \delt{a}{i}{j}  \wedge  \paths{v}{j}{k+1})) \label{suf2s}
\end{equation}
Finally, the complexity of the suffix model is reduced to ${\mathcal{O}}(\sigma\cdot k^2)$ instead of ${\mathcal{O}}(\sigma\cdot k^3)$.

For the hybrid model, construction is similar as before, but Constraints~(\ref{hyb1}) and (\ref{hyb2}) are simplified consequently: the index $i$ is removed from $\bigvee$ and $\bigwedge$, and changed to $k+1$ in paths and transitions:
  \begin{IEEEeqnarray}{c}
\bigvee_{j \in K} \paths{u}{1}{j} \wedge \paths{v}{j}{k+1}\\
\bigwedge_{j \in K} (\neg \paths{u}{1}{j} \vee \neg \paths{v}{j}{k+1}) \label{hyb1s}  \label{hyb2s}
\end{IEEEeqnarray}
The hybrid model has now the complexity ${\mathcal{O}}(\sigma\cdot k^2)$.

\subsection{Model specific refinements\label{sec_msref}}

The models presented thus far allow us to generate $(k+1)$\_NFA, but they do not guarantee that the obtained automaton is reducible to a $k$\_NFA. Hence, they do not guarantee obtaining the $(k+1)^\star$\_NFA. To improve the number of times we succeed, we now refine again the inference models. However, these refinements are not applicable to each previously presented model, but only to the prefix model of Sect.~\ref{sec_ref}, which we will call $P_{k+1}$.

We now add some constraints to obtain the $\PM{k+1}^R$ model.
Let $\fs{i}$ be Boolean variables meaning that $i$ is a possible final state of the $k$\_NFA:
\begin{itemize}
 \item No negative word finishes in a final state of the $k$\_NFA:
  \begin{equation}
  \bigwedge_{i \in K}\bigwedge_{w \in S^-} (\paths{w}{1}{i} \rightarrow \neg  \fs{i}).\label{ref1star}
\end{equation}
 \item Each final state of the $k$\_NFA is reached by at least one positive word $w$ ($w=va$, $v \in Pref(S)$, and $a \in \Sigma$). The paths from state $q_1$ to $q_{k+1}$ (the unique final state of the $(k+1)$\_NFA) and from $q_1$ to $q_i$ (final state of the $k$\_NFA) are the same except for the last transition:
  \begin{equation}
  \bigwedge_{i \in K} (\fs{i} \rightarrow
  \bigvee_{w \in S^+} \bigvee_{j \in K}(\paths{v}{1}{j} \wedge \delt{a}{j}{i} \wedge \delt{a}{j}{k+1})).\label{ref2star}
  \end{equation}
 \item Each positive word finishes at least in one final state of the $k$\_NFA:
  \begin{equation}
  \bigwedge_{w \in S^+} \bigvee_{i \in K}(\paths{w}{1}{i} \wedge \fs{i}).\label{ref3star}
  \end{equation}
\end{itemize}
\noindent
Note that this refinement, which generates $(k+1)_R$\_NFA, is only worth for the prefix models. For the other models, we still have to compute the suffixes for each pair of states, and we come back to the complexity of the $k$\_NFA models.

\section{A reduction algorithm: from $(k+1)$\_NFA to $k$\_NFA}
\label{sec_algo}

Let us consider we have a $(k+1)$\_NFA of $F_4$. 
Then, each transition $\delt{a}{i}{k+1}$ can be removed, and each state $j$ such that there is $\delt{a}{i}{j}$ must be considered as ``possibly'' final.
In the worst case, it means $2^k$ possibilities of $k$\_NFA (each of the $k$ states is either final or not). Each of these NFAs must be tested on $S$ to be validated. This is tractable, but we now present a simpler reduction algorithm.

Given a $(k+1)$\_NFA, we can try to reduce it to a $k$\_NFA with the following procedure:
\begin{enumerate}
 \item Based on the variables $\paths{w}{1}{i}$ for $i \in K$ and $w \in \Sm$, determine the set of candidate final states as $Q^A \setminus \{q_j\}$, where $j \in K$ and $\paths{w}{1}{j} = \texttt{true}$. This means that states in which any negative example can be reached cannot be final. 
 \item If the set of candidate final states is empty, the algorithm terminates, and the $k$\_NFA cannot be obtained from the given $(k+1)$\_NFA.
 \item Otherwise, given a non-empty set of candidate final states, for each word $w \in \Sp$ test whether the word can be reached in any candidate state by investigating variables $\paths{w}{1}{i}$, for these states. 
 \item If the test in the previous step returns a negative result for some word $w \in \Sp$ (i.e., the word cannot be reached in any of the candidate final states), the algorithm terminates, and the $k$\_NFA cannot be obtained from the given $(k+1)$\_NFA. 
 \item Otherwise, the $k$\_NFA can be obtained by removing the transitions leading to state $(k+1)$ and setting all candidate states to be final. 
\end{enumerate}

The worst-case complexity of the algorithm is in $O(k\cdot |S|)$, since Step 1 requires $k\cdot |\Sm|$ tests and Step 3 requires at most $k\cdot|\Sp|$ tests. Note also, that for a $(k+1)^\star$\_NFA obtained by $\PM{k+1}^R$ model the algorithm will always succeed. 

\smallskip
\noindent\textbf{Example 4}: Consider the $(k+1)^\star$\_NFA shown in Fig.~\ref{fig:examplea}. Step 1 determines the singleton set of candidate final states $\{q_2\}$ (all words $w \in \Sm$ can be reached in either $q_1$ or $q_3$, so they are excluded). Since the set is not empty, we move to step 3, and test whether each word $w \in \Sp$ is reachable in state $q_2$. Hence, we terminate with a $k$\_NFA as shown in Fig.~\ref{fig:examplea}.

\begin{figure}[!ht]
 \centering 
 \includegraphics[width=0.5\columnwidth]{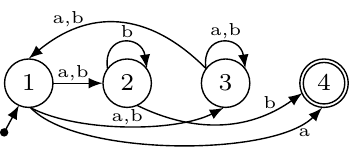}
 \caption{Example $(k+1)$\_NFA for sample $S = (\{a, ab, abba, baa\}$, $\{aab, b, ba, bab\})$}\label{fig:examplec}
\end{figure}

\smallskip
\noindent\textbf{Example 5}:
Consider now a $(k+1)$\_NFA shown in Fig.~\ref{fig:examplec} obtained for the sample $S$ from Example~1. With only one additional transition, $\delt{b}{1}{2}$, the $k$\_NFA cannot be built, since after step 1, the set of candidate states is empty.

\begin{figure}[!h]
 \centering 
 \includegraphics[width=0.5\columnwidth]{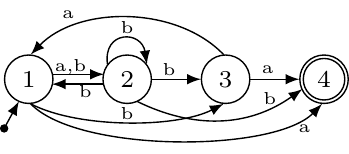}
 \caption{Example $(k+1)$\_NFA for sample $S = (\{ab, abba, ba, baa\}$, $\{aa, aab, b, bab\})$}\label{fig:exampled}
\end{figure}

\noindent\textbf{Example 6}:
Consider now a $(k+1)$\_NFA shown in Fig.~\ref{fig:exampled} obtained for the sample $S = (\Sp, \Sm)$ such that $\Sp = \{ab, abba, ba, baa\}$ and $\Sm = \{aa, aab, b, bab\}$. After step 1 the set of candidate states is $\{q_1\}$ (words $aa$ and $aab$ are not reachable at all; words $b$ and $bab$ exclude states $q_2$ and $q_3$). The tests in step 3 for words $ab, abba$, and $ba$ succeed, but it turns out that word $baa$ can only be reached in state $q_2$ which is not a candidate final state. Hence, we conclude that the $k$\_NFA cannot be built from the given $(k+1)$\_NFA.

Let us view Examples 4--6 from the perspective of constraints (\ref{ref1star})--(\ref{ref3star}). In Example~3 it is easy to verify that all three constraints are satisfied. In Example~4, after evaluating Constraint (\ref{ref1star}), we set all $f_i^k$ variables to \texttt{false}, thus satisfying also Constraint (\ref{ref2star}). However, we are then unable to satisfy Constraint (\ref{ref3star}). A similar observation is made for Example 5, even though we do not initially set all $f_i^k$ variables to \texttt{false}. In this case, constraint (\ref{ref3star}) cannot be satisfied for word $baa$.

\section{Experimentation}
\label{sec-expe}

\subsection{Models for $k$\_NFA vs. $(k+1)$\_NFA inference}

The algorithms were implemented in Python using libraries such as PySAT~\cite{pysat}. The experiments were carried out on a computing cluster with Intel-E5-2695 CPUs, and a fixed limit of 10 GB of memory. Running times were limited to 15 minutes, including model generation and solving time. We used the Glucose~\cite{glucose} SAT solver with default options.
	
In Tables \ref{table:comp} and \ref{table:summary} we compare the different models previously presented for $k$\_NFA and $(k+1)$\_NFA. Note that $\ILS{k}{init}{f}$ is used with 3 different initial splits: random split ($\ILS{k}{r}{f}$), split based on best prefix ($\ILS{k}{\PstarM{k}}{f}$), and split based on best suffix ($\ILS{k}{\SstarM{k}}{f}$). These experiments were carried out on state-of-the-art instances, described in~\cite{ICTAI2021}. These instances can be divided into three categories corresponding to the sizes of the alphabet (2, 5, and 10). The number of positive and negative words are the same in each instance and vary from 10 to 100 in increments of 10 for each category. There are thus 30 instances in total. The value of $k$ we used for the experiments is the best known lower bound for each instance found in various papers.

For each type of model, the $k$\_NFA and $(k+1)$\_NFA approaches were tested. The columns of Table~\ref{table:comp} and Table~\ref{table:summary} show the average number of variables and clauses as well as the number of resolved instances, and among the solved instances, the number of SAT instances. The column Ctime shows the cumulative execution time (generation of the model and solving time). Since the maximum execution time is 900 seconds, each category can consume at most 9000 seconds and thus, 27000 seconds for all the instances.

\begin{table*}
\centering
\caption{Comparison of the different models for $k$\_NFA and $(k+1)$\_NFA. The tested instances are divided into three categories according to their alphabet size. Each category contains 10 instances ranging from 20 to 200 words.}\label{table:comp}				
\begin{tabular}{|l|l||r|r|r|r|r||r|r|r|r|r||r|r|r|r|r|}																																	
\cline{3-17}																																	
\multicolumn{2}{c||}{}			&	\multicolumn{5}{c||}{\textbf{Size of the alphabet = 2}}									&	\multicolumn{5}{c||}{\textbf{Size of the alphabet = 5}}									&	\multicolumn{5}{c|}{\textbf{Size of the alphabet = 10}}									\\ \cline{3-17}
\multicolumn{2}{c||}{}			&	\multicolumn{1}{c|}{\textbf{\textit{var.}}}	&	\multicolumn{1}{c|}{\textbf{\textit{cl.}}}	&	\multicolumn{1}{c|}{\rotatebox{90}{\textbf{\textit{solved}}~}}	&	\multicolumn{1}{c|}{\rotatebox{90}{\textbf{\textit{SAT}}~}}	&	\multicolumn{1}{c||}{\textbf{\textit{Ctime}}~}	&	\multicolumn{1}{c|}{\textbf{\textit{var.}}}	&	\multicolumn{1}{c|}{\textbf{\textit{cl.}}}	&	\multicolumn{1}{c|}{\rotatebox{90}{\textbf{\textit{solved}}~}}	&	\multicolumn{1}{c|}{\rotatebox{90}{\textbf{\textit{SAT}}~}}	&	\multicolumn{1}{c||}{\textbf{\textit{Ctime}}~} &
\multicolumn{1}{c|}{\textbf{\textit{var.}}}	&	\multicolumn{1}{c|}{\textbf{\textit{cl.}}}	&	\multicolumn{1}{c|}{\rotatebox{90}{\textbf{\textit{solved}}~}}	&	\multicolumn{1}{c|}{\rotatebox{90}{\textbf{\textit{SAT}}~}}	&	\multicolumn{1}{c|}{\textbf{\textit{Ctime}}~}\\ \hline
\multirow{2}{*}{$\PM{k}$}	&	$k$	&	18 049	&	66 282	&	2	&	1	&	7 204	&	19 797	&	70 949	&	4	&	3	&	5 930	&	10 823	&	37 208	&	6	&	5	&	3 638	\\ 
	&	$k+1$	&	19 721	&	72 214	&	3	&	2	&	6 425	&	22 418	&	80 100	&	5	&	5	&	4 828	&	12 836	&	43 852	&	\textbf{7}	&	7	&	2 836	\\ \hline
\multirow{2}{*}{$\SM{k}$}	&	$k$	&	162 589	&	600 680	&	2	&	1	&	7 203	&	130 589	&	471 852	&	2	&	2	&	7 209	&	51 103	&	177 707	&	6	&	5	&	3 618	\\ 
	&	$k+1$	&	18 447	&	67 607	&	3	&	2	&	6 309	&	19 354	&	69 095	&	4	&	4	&	5 463	&	10 760	&	36 641	&	6	&	6	&	3 613	\\ \hline
\multirow{2}{*}{$\ILS{k}{r}{f}$}	&	$k$	&	24 586	&	94 912	&	2	&	1	&	7 220	&	22 073	&	82 352	&	4	&	3	&	5 625	&	10 659	&	38 354	&	5	&	4	&	4 569	\\ 
	&	$k+1$	&	7 718	&	28 059	&	\textbf{4}	&	3	&	5 765	&	12 987	&	46 119	&	5	&	5	&	4 551	&	8 971	&	30 486	&	\textbf{7}	&	7	&	2 767	\\ \hline
\multirow{2}{*}{$\ILS{k}{\PstarM{k}}{f}$}	&	$k$	&	24 166	&	93 393	&	2	&	1	&	7 219	&	22 045	&	82 305	&	4	&	3	&	5 594	&	10 583	&	38 080	&	5	&	4	&	4 593	\\ 
	&	$k+1$	&	7 834	&	28 485	&	3	&	2	&	6 328	&	12 943	&	45 961	&	5	&	5	&	4 564	&	9 051	&	30 760	&	\textbf{7}	&	7	&	2 782	\\ \hline
\multirow{2}{*}{$\ILS{k}{\SstarM{k}}{f}$}	&	$k$	&	15 781	&	62 423	&	2	&	1	&	7 220	&	17 383	&	65 431	&	4	&	3	&	5 596	&	9 280	&	33 536	&	5	&	4	&	4 595	\\ 
	&	$k+1$	&	12 891	&	47 189	&	\textbf{4}	&	3	&	5 468	&	17 621	&	62 884	&	5	&	5	&	4 892	&	9 978	&	34 003	&	\textbf{7}	&	7	&	3 015	\\ \hline
\multirow{2}{*}{$\PstarM{k}$}	&	$k$	&	107 425	&	401 231	&	2	&	1	&	7 204	&	100 673	&	366 869	&	3	&	3	&	6 392	&	39 232	&	138 115	&	6	&	5	&	4 762	\\ 
	&	$k+1$	&	12 402	&	45 381	&	3	&	2	&	6 305	&	15 202	&	54 167	&	5	&	5	&	4 972	&	8 625	&	29 271	&	6	&	6	&	3 612	\\ \hline
\multirow{2}{*}{$\SstarM{k}$}	&	$k$	&	15 811	&	62 503	&	3	&	2	&	6 307	&	18 062	&	67 697	&	4	&	3	&	5 669	&	9 890	&	35 495	&	5	&	4	&	4 559	\\ 
	&	$k+1$	&	12 962	&	47 384	&	3	&	2	&	6 315	&	17 646	&	62 969	&	\textbf{6}	&	6	&	3 749	&	9 961	&	33 949	&	\textbf{7}	&	7	&	2 747	\\ \hline
\end{tabular}						
\end{table*}																																	

\begin{table}														\centering		
\caption{Summary of the comparison of the different models for $k$\_NFA and $(k+1)$\_NFA.}	\label{table:summary}													
\begin{tabular}{|l|l||r|r|r|r|r|}																
\cline{3-7}																
\multicolumn{2}{c||}{}			&	\multicolumn{1}{c|}{\textbf{var.}}	&	\multicolumn{1}{c|}{\textbf{cl.}}	&	\multicolumn{1}{c|}{\rotatebox{90}{\textbf{solved}~}}	&	\multicolumn{1}{c|}{\rotatebox{90}{\textbf{SAT}~}}	&	\multicolumn{1}{c|}{\textbf{Ctime}~}			\\	\hline
\multirow{2}{*}{$\PM{k}$}	&	$k$	&	16 223	&	58 146	&	12	&	9	&	16 772			\\	
	&	$k+1$	&	18 325	&	65 389	&	15	&	14	&	14 089			\\	\hline
\multirow{2}{*}{$\SM{k}$}	&	$k$	&	114 760	&	416 747	&	10	&	8	&	18 031			\\	
	&	$k+1$	&	16 187	&	57 781	&	13	&	12	&	15 385			\\	\hline
\multirow{2}{*}{$\ILS{k}{r}{f}$}	&	$k$	&	19 106	&	71 873	&	11	&	8	&	17 413			\\	
	&	$k+1$	&	9 892	&	34 888	&	\textbf{16}	&	15	&	13 083			\\	\hline
\multirow{2}{*}{$\ILS{k}{\PstarM{k}}{f}$}	&	$k$	&	18 931	&	71 259	&	11	&	8	&	17 406			\\	
	&	$k+1$	&	9 943	&	35 069	&	15	&	14	&	13 674			\\	\hline
\multirow{2}{*}{$\ILS{k}{\SstarM{k}}{f}$}	&	$k$	&	14 148	&	53 797	&	11	&	8	&	17 411			\\	
	&	$k+1$	&	13 496	&	48 025	&	\textbf{16}	&	15	&	13 374			\\	\hline
\multirow{2}{*}{$\PstarM{k}$}	&	$k$	&	82 443	&	302 072	&	11	&	9	&	18 358			\\	
	&	$k+1$	&	12 077	&	42 940	&	14	&	13	&	14 889			\\	\hline
\multirow{2}{*}{$\SstarM{k}$}	&	$k$	&	14 587	&	55 232	&	12	&	9	&	16 535			\\	
	&	$k+1$	&	13 523	&	48 101	&	\textbf{16}	&	15	&	\textbf{12 810}			\\	\hline
\end{tabular}																
\end{table}															
Based on Table \ref{table:comp}, we see that the size of the alphabet $\Sigma$ does not influence much the size of the instances: $|\Sigma|$ does not appear in the complexity which is upper bounded by $\sigma$; but in practice, the number of prefixes (or suffixes) is related to $|\Sigma|$ (the probability to have common prefixes is higher when $|\Sigma|$ is small). The size of the instances significantly depends on the models, and on the samples (number of prefixes or suffixes). $\SM{k}$ and $\PstarM{k}$ generate the largest models, and the difference with other models is significant.

In Table \ref{table:summary}, we can notice that either $\ILSs{k}{r}$, $\ILSs{k}{\SstarM{k}}$, or $\SstarM{k}$ are able to solve more than half of the instances of the tested samples. $\SstarM{k}$ is the fastest model for the total generation time and solving time. It enables us to save 3725 seconds for $(k+1)$\_NFA which can then be used for the reduction to a $k$\_NFA.

On average, $(k+1)$ models are smaller than $k$\_NFA models in terms of the number of variables. Whatever the model, more $(k+1)$\_NFA can be inferred than $k$\_NFA. This is due to both smaller instances and more constrained instances. Although $(k+1)$ models can solve more instances, the cumulative time is always smaller than for $k$\_NFA: this means that solved instances are also solved faster. $\ILS{k}{init}{f}$ is not very sensitive to the initial splitting of words. However, it seems that the random initialization enables us to obtain smaller SAT instances. It seems that better initialization such as with best prefixes or best suffixes stay stuck in local minima, close to the initialization.

The statistical analysis of the $k$\_NFA vs. $(k+1)$\_NFA models shows no statistically significant differences in the execution times if we assume the execution time of 900 seconds for the unsolved instances. However, when we focus only on the solved instances, we get statistically significant differences based on the Kruskal-Wallis test ($p$-value = $1.4 \cdot 10^{-6}$). The post-hoc Dunn's analysis shows that the significant differences are between all ILS-based models and the rest of the models (in favor of the other models, which have lower mean ranks). There are no significant differences between $k$\_NFA and $(k+1)$\_NFA for the same model type.

\subsection{$(k+1)$\_NFA to $k$\_NFA reduction algorithm}

The results shown in Tables \ref{table:comp} and \ref{table:summary} were obtained for the base and refined models, excluding the refinements of Sect.~\ref{sec_msref}. Consequently, the vast majority of $(k+1)$\_NFA could not be reduced to $k$\_NFA. In the second experiment, all instances have been tested with the $\PM{k}$ model for $k$\_NFA, $(k+1)$\_NFA and reducible $(k+1)_R$\_NFA with a maximum running time of 15 minutes (900 seconds). 

Table \ref{tab:kp1tok} details the results for instances solved and proved SAT. For each instance, the $k$ column provides three values corresponding to the sizes of the $k$\_NFA, $(k+1)$\_NFA, and reducible $(k+1)_R$\_NFA, respectively. Columns $var.$ and $cl.$ represent the number of variables and clauses. Column $Stime$ is the generation and solving time, while $Rtime$ is the running time of the reduction algorithm. Finally, $Ttime$ is the total running time.

For each instance the best time to obtain a $k$-state NFA (either directly or through reduction) is bolded. A star ($\star$) is added when the reduction of the reducible $(k+1)_R$\_NFA permits to find a $k$\_NFA whereas it is not possible directly using the model for $k$\_NFA. When it takes longer to reduce $(k+1)_R$\_NFA to a $k$\_NFA than to find the $k$\_NFA, a bullet ($\bullet$) is added. When the running time exceeded 15 minutes, nothing is written. Clearly, reduction time is only provided for $(k+1)$\_NFA and reducible $(k+1)_R$\_NFA.

\begin{table}[hbt!]
\centering
\caption{Detailed results for $\PM{k}$ model for $k$\_NFA, $(k+1)$\_NFA and reducible $(k+1)_R$\_NFA. The results include only solved and proved SAT instances.}\label{tab:kp1tok}
\begin{tabular}{|l|l|r|r|r|r|r|}
\hline
\multicolumn{1}{|c|}{\textbf{Instance}} & \multicolumn{1}{c|}{\textbf{$k$}} & \multicolumn{1}{c|}{\textbf{var.}} & \multicolumn{1}{c|}{\textbf{cl.}} & \multicolumn{1}{c|}{\textbf{Stime}} & \multicolumn{1}{c|}{\textbf{Rtime}} & \multicolumn{1}{c|}{\textbf{Ttime}}\\ \hline
 & 4 & 1 276 & 4 250& 1.67 & & 1.67\\ 
 & 5 & 1 550 & 5 128& 1.88 & 0.01 & 1.89\\ 
\multirow{-3}{*}{st-2-10}  & 5 R& 1754 & 5942& 0.96 & 0.01 & \textbf{0.97}\\ \hline
 & 7 & 6 461 & 23 113& & &\\ 
 & 8 & 7 232 & 25 806& 121.66  & 0.01 & 121.68 \\ 
\multirow{-3}{*}{st-2-20}  & 8 R & 8 359 & 30 313& 185.44 & 0.02 & \textbf{185.46$\star$} \\ \hline
 & 3 & 1 098 & 3 445& 1.82 & & 1.82\\ 
 & 4 & 1 440 & 4 455& 1.77 & 0.00 & 1.78\\ 
\multirow{-3}{*}{st-5-10}  & 4 R & 1 563 & 4 948& 1.07 & 0.01 & \textbf{1.07}\\ \hline
 & 4 & 4 024 & 13 464& 1.99 & & 1.99\\ 
 & 5 & 4 950 & 16 465& 1.85 & 0.00 & 1.85\\ 
\multirow{-3}{*}{st-5-20}  & 5 R & 5 354 & 18 089& 1.06 & 0.01 & \textbf{1.07}\\ \hline
 & 5 & 8 020 & 27 720& 4.84 & & 4.84\\ 
 & 6 & 9 473 & 32 593& 1.66 & 0.01 & \textbf{1.67}\\ 
\multirow{-3}{*}{st-5-30}  & 6 R & 10 373& 36 228& 2.23 & 0.05 & 2.28\\ \hline
 & 5 & 9 390 & 32 550& 19.77& & \textbf{19.77$\bullet$}  \\ 
 & 6 & 11052& 38157& 11.58& 0.01 & 11.59  \\ 
\multirow{-3}{*}{st-5-40}  & 6 R & 12257& 43002& 20.62& 0.02 & 20.63  \\ \hline
 & 6 & 15 564& 55 100& & &\\ 
 & 7 & 17 836& 62 955& 354.36  & 0.02 & 354.38 \\ 
\multirow{-3}{*}{st-5-50}  & 7 R & 20 706& 74 327& & &\\ \hline
 & 3 & 831  & 2 431& 1.71 & & 1.71\\ 
 & 4 & 1 104 & 3 118& 1.54 & 0.00 & 1.54\\ 
\multirow{-3}{*}{ww-10-10} & 4 R & 1 227 & 3 611& 1.27 & 0.00 & \textbf{1.27}\\ \hline
 & 3 & 2 613 & 8 255& 1.65 & & 1.65\\ 
 & 4 & 3 440 & 10 730& 1.63 & 0.00 & 1.63\\ 
\multirow{-3}{*}{ww-10-20} & 4 R & 3 683 & 11 713& 1.01 & 0.06 & \textbf{1.07}\\ \hline
 & 4 & 1 443 & 4 485& 1.60 & & 1.60\\ 
 & 5 & 1 840 & 5 550& 1.58 & 0.00 & 1.58\\ 
\multirow{-3}{*}{ww-10-30} & 5 R & 3 479 & 11 459& 1.06 & 0.01 & \textbf{1.06}\\ \hline
 & 4 & 3 624 & 11 900& 1.39 & & 1.39\\ 
 & 5 & 4 375 & 14 145& 1.53 & 0.01 & 1.53\\ 
\multirow{-3}{*}{ww-10-40} & 5 R & 5 179 & 17 389& 1.00 & 0.03 & \textbf{1.03}\\ \hline
 & 4 & 5 364 & 17 850& 1.67 & & 1.67\\ 
 & 5 & 6 500 & 21 390& 2.44 & 0.01 & 2.45\\ 
\multirow{-3}{*}{ww-10-50} & 5 R & 7 504 & 25 444& 1.59 & 0.01 & \textbf{1.60}\\ \hline
 & 5 & 10 995& 37 800& 27.18& & 27.18  \\ 
 & 6 & 12 888& 44 018& 3.23 & 0.01 & 3.24\\ 
\multirow{-3}{*}{ww-10-60} & 6 R & 14 693& 51 283& 12.27& 0.03 & \textbf{12.30}  \\ \hline
 & 5 & 14 525& 50 190& & &\\ 
 & 6 & 17 064& 58 654& 121.27  & 0.01 & 121.28 \\ 
\multirow{-3}{*}{ww-10-70} & 6 R & 19 169& 67 129& 349.95  & 0.02 & \textbf{349.98$\star$} \\ \hline
\end{tabular}
\end{table}

We can observe in Table \ref{tab:kp1tok} that reducible $(k+1)_R$\_NFA obtains the best total time for 12 of the 14 instances. Moreover, for instances \texttt{st-2-20} and \texttt{ww-10-70} NFA is found by reduction of the $(k+1)_R$\_NFA whereas no NFA is found by $k$\_NFA. It proves that it is worth using reducible $(k+1)_R$\_NFA rather than $k$\_NFA.
We can note that only instances \texttt{st-5-30} and \texttt{ww-10-30} can be reduced from  $(k+1)$\_NFA to $k$\_NFA, and the total running time is better only once than for reducible $(k+1)_R$\_NFA.

The number of variables and clauses of reducible $(k+1)_R$\_NFA is bigger than for $(k+1)$\_NFA but it does not impact the efficiency. Moreover, the reduction time is irrelevant with respect to the solving time.

\section{Conclusion}
\label{sec_conclusion}

Grammatical inference consists in learning a formal
grammar, in our case as an NFA. We have presented how we can benefit from a very simple property to refine and improve some already refined models from~\cite{ICTAI2021}. The property says that if there is a $k$\_NFA, there is also a $(k+1)$\_NFA. This property has been strengthened to bring us closer to the $(k+1)^\star$\_NFA that one could build manually.

Thanks to the proposed refinements, we managed to obtain smaller models, with a lower complexity decreasing from ${\mathcal{O}}(\sigma\cdot k^3)$ to ${\mathcal{O}}(\sigma\cdot k^2)$. With some experiments, we have shown that the $(k+1)$\_NFA are easier and faster to infer than the $k$\_NFA. We have also presented an algorithm to reduce $(k+1)$\_NFA to $k$\_NFA. Although the $(k+1)$ models are closer to the $(k+1)^\star$\_NFA obtainable with the simple algorithm given in this paper, the reduction algorithm does not always succeed. We have shown that given some model specific refinements, we can keep the low complexity of the model and infer $(k+1)_R$\_NFA that can be very quickly reduced to $k$\_NFA.

In the future, we plan to refine again the $(k+1)$\_NFA models for the reduction algorithm to succeed more often. For example, we want to add constraints such as the ones for the $P_k$ model. We also plan to integrate symmetry breaking techniques to reduce the initial search space. But to this end, we will have to be very cautious and verify that symmetry breaking constraints are compatible with the $(k+1)$\_NFA models.

\bibliographystyle{IEEEtran}
\bibliography{biblio}

\end{document}